\documentclass[runningheads]{llncs}

\usepackage[T1]{fontenc}
\usepackage{graphicx}

\usepackage{amsmath}
\usepackage{xcolor}
\usepackage{listings}
\usepackage{makecell}
\usepackage{multicol}
\usepackage{multirow}
\usepackage{wrapfig}
\usepackage{marvosym}
\usepackage{cite}
\usepackage[hidelinks]{hyperref}
\usepackage{orcidlink}

\begin{document}
\title{Towards Scalable GPU-Accelerated SNN Training \\via Temporal Fusion}

\author{
    Yanchen Li\orcidlink{0009-0002-2237-0451} \and
    Jiachun Li\orcidlink{0009-0004-4182-9664}\and
    Kebin Sun\orcidlink{0009-0008-9213-7835} \and
    Luziwei Leng\orcidlink{0000-0002-9344-8589} \and
    Ran Cheng\textsuperscript{(\Letter)}\orcidlink{0000-0001-9410-8263}
}
\authorrunning{Y. Li \emph{et al.}}
\institute{Department of Computer Science and Engineering, Southern University of Science and Technology, Shenzhen, China \\
\email{ranchengcn@gmail.com}}

\maketitle

\vskip -1.5em
\renewcommand\thefootnote{}
\footnotetext{This work was supported in part by Guangdong Natural Science Funds for Distinguished Young Scholar under Grant 2024B1515020019.}
\footnotetext{Yanchen Li and Jiachun Li contributed equally to this work.}
\footnotetext{Corresponding author: Ran Cheng (\texttt{ranchengcn@gmail.com})}

\begin{abstract}
Drawing on the intricate structures of the brain, Spiking Neural Networks (SNNs) emerge as a transformative development in artificial intelligence, closely emulating the complex dynamics of biological neural networks. 
While SNNs show promising efficiency on specialized sparse-computational hardware, their practical training often relies on conventional GPUs. 
This reliance frequently leads to extended computation times when contrasted with traditional Artificial Neural Networks (ANNs), presenting significant hurdles for advancing SNN research.
To navigate this challenge, we present a novel temporal fusion method, specifically designed to expedite the propagation dynamics of SNNs on GPU platforms, which serves as an enhancement to the current significant approaches for handling deep learning tasks with SNNs.
This method underwent thorough validation through extensive experiments in both authentic training scenarios and idealized conditions, confirming its efficacy and adaptability for single and multi-GPU systems.  
Benchmarked against various existing SNN libraries/implementations, our method achieved accelerations ranging from {\(5\times\) to \(40\times\)} on NVIDIA A100 GPUs. Publicly available experimental codes can be found at \url{https://github.com/EMI-Group/snn-temporal-fusion}.

\end{abstract}
\section{Introduction}
Deep learning has firmly entrenched itself as a transformative field in artificial intelligence. 
Python, as a preferred programming language, has propelled this transformation. 
Notably, by leveraging the robust GPU computing supports of the Compute Unified Device Architecture (CUDA)~\cite{DBLP:conf/isbi/Luebke08}, PyTorch~\cite{DBLP:conf/nips/PaszkeGMLBCKLGA19} stands out in the Python ecosystem, offering a robust platform for researchers working on traditional Artificial Neural Network (ANN) structures.

As the deep learning landscape evolves, the integration of biological neural dynamics, manifested in the form of Spiking Neural Networks (SNNs), has emerged as a promising research direction. 
Presently, the spiking attribute is primarily viewed as a module or characteristic that can be integrated into existing ANNs. 
This perspective is largely influenced by studies focused on converting ANNs into SNNs~\cite{DBLP:conf/ijcai/DingY0H21,DBLP:conf/icml/LiD0GG21,10.3389/fnins.2017.00682}, as well as training SNNs directly with ANN methods~\cite{DBLP:conf/nips/LiGZDHG21,10.1007/978-3-030-30487-4_54,DBLP:conf/aaai/WuDLZ0S19}. 
Consequently, ANNs serve as foundational pillars for contemporary SNNs. 
Reflecting this trend, numerous SNN libraries have been developed atop the PyTorch framework, each distinguished by its unique focus~\cite{DBLP:journals/pieee/EshraghianWNWLDBJL23,doi:10.1126/sciadv.adi1480,DBLP:journals/fini/HazanSKPSSK18,norse2021}.

Despite the promising features of SNNs, a significant challenge arises with the incorporation of the \emph{temporal dimension}, which often leads to slower training speeds. 
Many researchers have sought to mitigate this by truncating the network's temporal length~\cite{DBLP:conf/eccv/ChowdhuryRR22,11.3389/fnins.2023.1047008,10.1007/978-3-031-44192-9_33}. While expediting training, this approach inadvertently suppresses its temporal characteristics and makes SNNs closely resemble ANNs~\cite{DBLP:conf/iclr/BuFDDY022,DBLP:conf/cvpr/0006S020,DBLP:journals/tnn/RathiR23}.
Moreover, the relatively slow execution speeds of SNNs on prevalent GPU platforms make it challenging to train and validate the models on large scales. 
Undoubtedly, such computational bottlenecks have impeded further exploration of the field.

To address these challenges, various \emph{upper-level} acceleration algorithms have been proposed, focusing on the integration of spiking neuron mechanisms with advanced training strategies~\cite{fang2023parallel,DBLP:conf/iccv/Meng0Y0LL23,DBLP:conf/nips/XiaoMZHL22}. 
In contrast, research on accelerating deep learning models has predominantly concentrated on \emph{lower-level} computational improvements. 
These involve parallel and distributed processing techniques applied to both data and models, which have primarily been implemented in ANNs to boost efficiency in solving deep learning tasks~\cite{DBLP:conf/nips/HuangCBFCCLNLWC19,DBLP:conf/nips/KrizhevskySH12,DBLP:journals/corr/abs-1909-08053}.
While these contributions are noteworthy, the lower-level computational infrastructures specifically designed for SNNs are yet to be further explored.

To bridge the gap, in this work, we seek to address the challenge of low-speed execution inherent in SNNs by introducing a theoretically sound solution, leveraging the temporal dynamics of SNNs during signal propagation.
This method effectively reduces the latency associated with the temporal aspects of processing, thereby facilitating the scalable training of SNNs across distributed GPU environments. 
Our contributions can be summarized as follows.
\begin{itemize}
    \item 
    We propose a novel \emph{temporal fusion} method specifically designed to expedite the training of SNNs. 
    Drawing inspiration from prior studies, our method decouples the spiking neuron model's propagation patterns along the temporal axis. 
    This strategy facilitates more efficient information flow and maintains computational precision without resorting to approximate substitutions, thereby ensuring that the model's outcomes are both simple and accurate.
         
    \item 
    We have extended the proposed temporal fusion method to leverage multiple GPUs, moving beyond traditional batch-based methods. 
    We adopt a pipeline parallelism framework, which is particularly suitable for the intrinsic temporal dynamics of spiking neurons, enabling the distribution of computational loads across GPUs in a temporal manner. 
    Theoretically, this design allows for scalable performance improvements that grow with the increase in time steps, offering significant acceleration benefits.
    
    \item 
    To empirically validate the acceleration capabilities of our method, we conducted experiments on deep learning tasks using widely adopted SNN architectures during both training and testing phases.
    Our implementation, developed on the PyTorch platform with CUDA optimization, is specifically engineered to reduce memory access latency, thereby enhancing efficiency across various temporal scales. 
    The results from our rigorous testing, devoid of extraneous computational burdens, demonstrate substantial performance improvements in both single and multi-GPU setups.
\end{itemize}

\section{Related Work}

\subsection{GPU Acceleration for Deep Learning}
In the realm of GPU acceleration for deep learning, NVIDIA's CUDA technology is particularly noteworthy for its high-level interface that enables direct interaction with GPUs. 
CUDA serves as a crucial software layer that allows developers to engage directly with the GPU's virtual instruction set and parallel computing components, thus facilitating efficient execution of kernel computations. 
This proximity to the hardware layer significantly enhances development flexibility and efficiency.

To further streamline the development process, CUDA is complemented by a suite of optimized standard routines.
A primary example is cuDNN~\cite{DBLP:journals/corr/ChetlurWVCTCS14}, which provides a comprehensive set of deep neural network operator implementations within the CUDA ecosystem, substantially simplifying the development of deep learning applications. 
These foundational technologies underpin many high-level deep learning frameworks, enabling robust and efficient experimentation.

There have also been significant advancements in parallel and distributed acceleration methods. 
Initial efforts, such as those by Krizhevsky \textit{et al.}~\cite{DBLP:conf/nips/KrizhevskySH12}, allocated neural network models across multiple GPUs, paving the way for processing large deep-learning models with multiple GPUs. 
Following this, more sophisticated distributed training methods have emerged.
For example, Megatron-LM~\cite{DBLP:journals/corr/abs-1909-08053} introduced an intra-layer model parallelism method for large language models, enhancing accuracy while maintaining performance. 
GPipe~\cite{DBLP:conf/nips/HuangCBFCCLNLWC19} explored serial operators in foundation models, implementing pipelined serial sub-layers and facilitating pipeline parallelism across multiple GPUs. 
These methods significantly boost model computation efficiency on supported platforms, with many integrated into existing deep learning frameworks to augment the efficiency of large-scale model development and research.

\subsection{Neuromorphic Computing Infrastructures}
In light of the integration of spiking neuron properties into deep learning, several infrastructures tailored for SNNs have emerged, each focusing on different facets of the domain. 
BindsNET~\cite{DBLP:journals/fini/HazanSKPSSK18} emphasizes applications in machine and reinforcement learning. 
Norse~\cite{norse2021} and snnTorch~\cite{DBLP:journals/pieee/EshraghianWNWLDBJL23} expand on SNNs within the PyTorch ecosystem, prioritizing comprehensive functional support alongside an extensive documentation suite, praised for its user-centric design. 
SpikingJelly~\cite{doi:10.1126/sciadv.adi1480}, on the other hand, accentuates algorithmic advancements in SNNs, offering two backends and two SNN propagation mechanisms, thereby fostering both algorithmic enhancements and performance optimizations. 

In addition, certain simulation frameworks for neural systems have inspired the fundamental design of SNNs.  
Libraries such as cuSNN \cite{DBLP:journals/pami/Paredes-VallesS20} and GeNN \cite{yavuz2016genn}, built directly on CUDA, aim to minimize runtime overhead through efficient, substrate-based designs. 
These frameworks facilitate a closer interaction with the hardware layer, enhancing the execution efficiency of SNN simulations.

\subsection{Spiking Neurons}
As the fundamental units of the brain, neurons exhibit unique information transfer properties. 
To emulate these intricate behaviours in computational models, various spiking neuron models have been proposed. 
Notably, the Spike Response Model (SRM)~\cite{gerstner_kistler_naud_paninski_2014} represents a broad category of spiking neurons, encompassing parameters like membrane potential decay, spike threshold, refractory period, etc.
Essentially, the membrane potential of a neuron undergoes continuous decay unless it receives an external stimulus. 
Upon receiving information, the potential increases until it hits a threshold, resulting in the generation of a spike, followed by an immediate potential drop. 

Spiking neuron models for deep learning are relatively complex in the early stages, e.g., Hodgkin-Huxley~\cite{https://doi.org/10.1113/jphysiol.1952.sp004717}, Izhikevich~\cite{Izhikevich20031569}. 
However, complex models, while accurate, often introduce implementation challenges. 
As the deep learning domain advances, the need for simplicity becomes paramount. 
Therefore, the Leaky-Integrate-and-Fire (LIF) model and its variants are preferred at present~\cite{DBLP:journals/ijon/ChowdhuryLR21,shaban2021adaptive,teeter2018generalized}. 
LIF emerges as a simplified version of SRM, preserving its core information transfer mechanisms while reducing the intricacies of neuronal information transmission. 
Recognizing the potential of this simplification, recent research has further refined the LIF model, yielding an iterable form~\cite{DBLP:conf/aaai/WuDLZ0S19}. 
This adaptation ensures that spiking neurons seamlessly integrate into the deep learning paradigm without compromising the salient features of SRM.

\section{Method}
We first analyze the properties of LIF neurons in forward and backward propagation, showcasing the potential for acceleration optimization. 
Then, we delve into the foundational aspects of the temporal fusion method and its theoretical implications when executed on a single GPU.
Finally, we explore the method's scalability across multiple GPUs.

\begin{figure*}[t]
    \centering
    \begin{minipage}{0.40\textwidth}
        \centering
        \includegraphics[width=\textwidth]{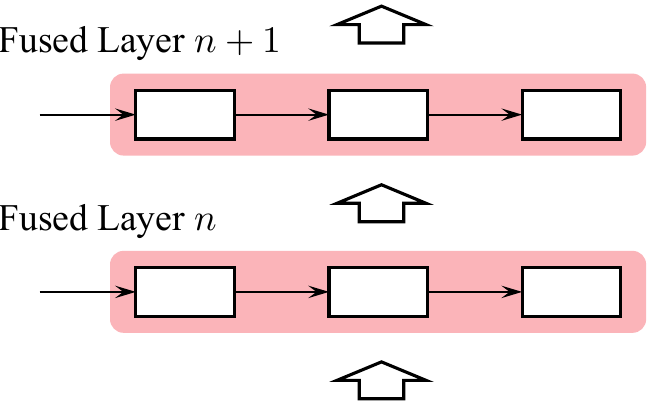}
        \newline \vskip 0.5em
        \footnotesize
        (a) Single GPU case
    \end{minipage}
    \begin{minipage}{0.01\textwidth}
        \tiny
        \textcolor{white}{1}
    \end{minipage}
    \begin{minipage}{0.57\textwidth}
        \centering
        \includegraphics[width=\textwidth]{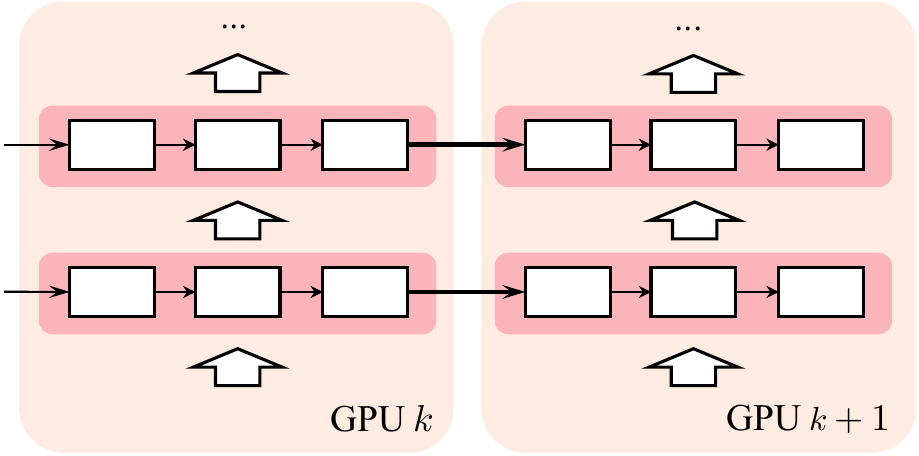}
        \newline \vskip -0.8em
        \footnotesize
        (b) Multiple GPUs case
    \end{minipage}
    \vskip 0em
    \caption{Schematics showcasing temporal fusion on both single and multi-GPU environments. The horizontal axis represents time step, while the vertical axis represents the hierarchical propagation of the networks. The temporal dimension is unfolded, with each square representing the neuron collection of a specific SNN layer at a given time step. }
    \label{fig:comparison}
    \vskip -1.5em
\end{figure*}

\subsection{Parallelism of LIF Spiking Neurons}
For a specific layer within the iterative LIF model of a SNN, let \(v_i^{(t)}\) denote the membrane potential of the \(i\)-th neuron at the \(t\)-th time step, and \(y_i^{(t)}\) represent its corresponding output. 
Initially, we assume the membrane potential of each neuron is \(v_i^{(0)} = V_\text{rest}\), where \(V_\text{rest}\) indicates the resting potential. 
Additionally, \(k_\tau\) is defined as the decay factor that influences the membrane potential's evolution over time.
Given these, the iterative dynamics of the LIF model for each neuron can be formulated as:
\begin{equation}
    v_i^{(t)} = k_{\tau} \cdot v_i^{(t-1)} \cdot \left(1 - y_i^{(t-1)}\right) + V_\text{rest} \cdot y_i^{(t-1)} + x_i^{(t)},
\label{eq:LIF}
\end{equation}
where the mechanism for spike generation is given by:
\begin{equation}
    y_i^{(t)} = H\!\!\left(v_i^{(t)} - V_\text{th}\right) =
    \begin{cases}
    1, & \text{if } v_i^{(t)} - V_\text{th} \geq 0, \\\\[-1.5ex]
    0, & \text{otherwise},
    \end{cases}
\label{eq:spike}
\end{equation}
with \(H\) denoting the spike activation function characteristic of spiking neurons, and \(x_i^{(t)}\) being the integrated output from the preceding layer's neurons.

Eqs.~\eqref{eq:LIF} and \eqref{eq:spike} elucidate the forward propagation mechanism of the iterative LIF model, providing a detailed account of the spiking process while preserving the essential attributes of spiking neurons. 
Consequently, the discussions and methodologies presented in this paper are based on these formulations.

The structure of the LIF model's forward propagation inherently supports its backward propagation, which is essential for direct SNN training. 
This backward propagation is expressed as:
\begin{align}
\nabla_{x_i^{(t)}}L &= k_\tau\cdot\nabla_{v_i^{(t+1)}}L\cdot[1-y_i^{(t)}-v_i^{(t)}\cdot\delta(v_i^{(t)})] + \nabla_{y_i^{(t)}}L\cdot\delta(v_i^{(t)}),
\label{eq:backLIF}
\end{align}
where \(L\) denotes the loss function. 
The spike function \(H\), being non-differentiable as indicated in Eq.~\eqref{eq:spike}, necessitates \(\delta\) as an approximate surrogate gradient function.

Notably, both forward and backward computations in the LIF model are conducted on an element-wise basis, as opposed to the fully connected and convolutional operations. 
This means that each neuron functions autonomously, without reliance on a specific computational sequence or interaction with others. 
Such autonomy is pivotal, as it allows for parallel processing and optimization, significantly boosting computational efficiency.

\begin{figure*}[!t]
    \centering
    \includegraphics[width=1\textwidth]{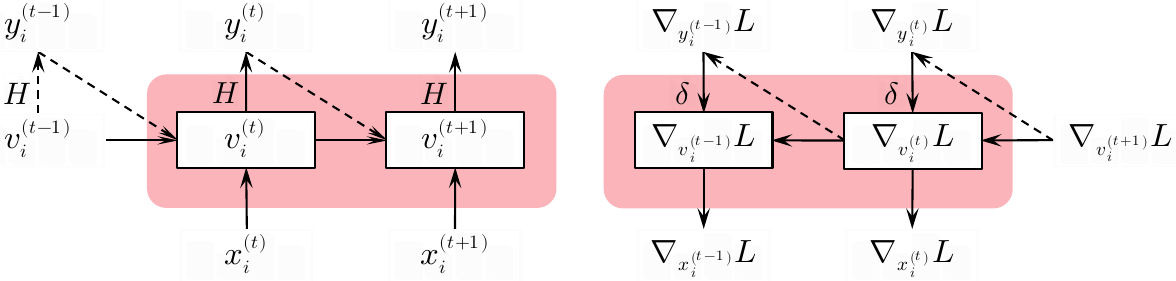}
    \vskip -0.5em
    \caption{Forward (see left) and backward (see right) propagation in a monolayer LIF network via temporal fusion. The horizontal axis indicates time-step-wise propagation, and the vertical axis shows layer-by-layer progression, conforming to Eqs.~\eqref{eq:LIF}, \eqref{eq:spike} and \eqref{eq:backLIF}. The red-shaded area delineates the operator fusion range within the GPU kernel, merging \(x_i^{(t)}\) and \(x_i^{(t+1)}\) (as well as \(\nabla_{y_i^{(t)}}L\) and \(\nabla_{y_i^{(t-1)}}L\)) for integrated GPU kernel processing, thereby minimizing the memory access overhead.}
    \label{fig:fused_forward_backward}
    \vskip -1em
\end{figure*}

\subsection{Temporal Fusion on a Single GPU}

As given by Eq.~\eqref{eq:LIF}, the dynamics of the membrane potential in SNNs are intricately tied to both temporal and spatial factors. 
This entails that the activity of any given neuron is predicated on its preceding state at the last time step, as well as the input it receives from neurons within the previous layer. 

Drawing inspiration from prior research~\cite{doi:10.1126/sciadv.adi1480}, an intriguing observation emerges: \emph{Time-step-wise inference is fundamentally equivalent to layer-wise inference, which occurs after the completion of all time steps in each layer of the computation.} 
This characteristic inherent to the latter opens a path for optimization across the temporal dimension, particularly when the initial states for all time steps are predetermined.

Notably, prioritizing the computation across each layer over all time steps yields two primary benefits. 
First, the element-wise computation inherent to the monolayer LIF model naturally facilitates parallelization at the neuron level, a feature that extends to temporal propagation.  
Second, executing computations across all time steps mandates the inclusion of data from all preceding time steps. 
Given that data for each time step retains a consistent format, strategic memory alignment post-concatenation can significantly reduce the overhead tied to batch memory operations, thereby elevating computational efficiency.

Building upon these insights, we introduce the temporal fusion method, which leverages the element-wise nature of spiking neuron layers. 
This method assigns each neuron's computation to an individual GPU thread, amalgamating memory operations for each layer across all time steps within the GPU kernel.  
Such fusions substantially reduce the overhead from multiple memory interactions and thus augment computational throughput. 
The overarching framework is illustrated in Fig.~\ref{fig:comparison}(a).

\begin{figure*}[!t]
    \centering
    \includegraphics[width=1\textwidth]{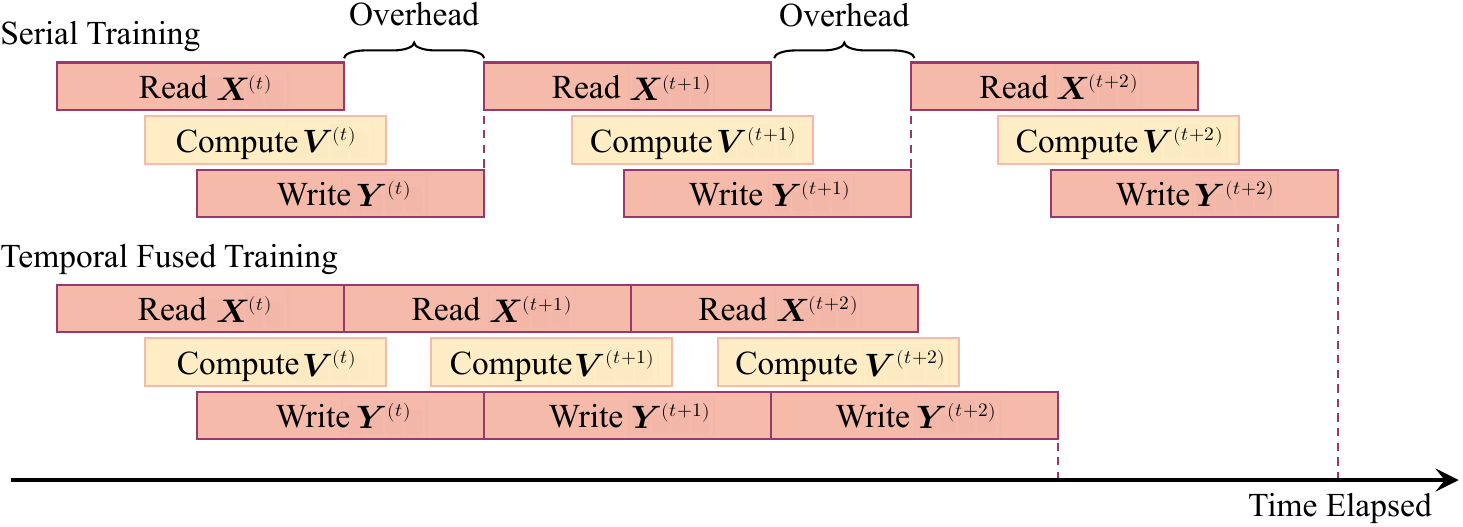}
    \vskip -1em
    \caption{A comparative analysis of the traditional serial method versus the temporal fusion method in SNN training.
     "Compute" refers to GPU kernel computations, while "read" and "write" pertain to memory operations. 
     \(\boldsymbol{X}^{(t)}\), \(\boldsymbol{V}^{(t)}\), and \(\boldsymbol{Y}^{(t)}\) denote the tensor representations of neuronal input, membrane potential, and output, respectively, conforming to the element-wise variables \(x_i^{(t)}\) and \(v_i^{(t)}\) in Eq.~\eqref{eq:LIF}, along with \(y_i^{(t)}\) in Eq.~\eqref{eq:spike}.}
    \label{fig:fused_analysis}
    \vskip -1em
\end{figure*}

For detailed insight, Fig.~\ref{fig:fused_forward_backward} illustrates the forward and backward propagation processes under the temporal fusion method, respectively. 
In these figures, each block containing neuronal information relies on preceding data within the unfolded temporal dimension, underscoring the interconnected nature of computations.

Theoretically, training with temporal fusion diverges from conventional serial training, which typically emphasizes sequential computation at the network layer. 
Instead, temporal fusion adopts a temporal-major order approach, computing all time steps within each layer to facilitate temporal operator fusion. 
Although reordering computations does not impact the end results, strategic utilization of this aspect can effectively minimize access losses. 
The contrast between serial and temporal fusion-based training is showcased in Fig.~\ref{fig:fused_analysis}, illustrating how serial training's frequent, minor memory operations incur linearly scaling overhead; contrastively, temporal fusion training sidesteps such overhead, offering increased benefits with a rising number of time steps.

\subsection{Temporal Fusion across Multiple GPUs}
As elaborated in the preceding subsection, the temporal fusion method enhances computational efficiency by minimizing memory access and leveraging operator fusion, even with small time steps. 
However, as time step sizes increase, a single GPU may encounter bottlenecks due to expanded temporal dimensions, potentially leading to storage constraints. 
The limited memory of a single GPU could necessitate halting computations or reverting to less efficient serial processing upon exceeding capacity. 
Furthermore, larger time steps contribute to increased computational latency. 
Utilizing multiple GPUs can mitigate these issues by distributing the computational workload. Consequently, we extend the temporal fusion method to a multi-GPU setup, as illustrated in Fig.~\ref{fig:comparison}(b).

In this expanded configuration, inter-GPU communication enables cross-device operator fusion and data exchange along the temporal dimension. 
Suppose \(k~(k \geq 1)\) identical GPUs are available. 
If the computational load for a specific SNN layer is enough to be divided into \(k\) segments along the temporal dimension on a single GPU, the completion time for \(k\) sub-tasks on these GPUs would be \(t^{(k)}\). 
Assuming the inter-GPU communication time is \(T_\text{c}\). Ideally, we have:
\begin{subequations}
    \begin{align}
        T_\text{s} &= k\cdot t^{(k)}, \\
        T_\text{m}^{(k)} &= (k-1)\cdot T_\text{c} + t^{(k)},
        \label{eq:multi}
    \end{align}
\end{subequations}
where \(T_\text{s}\) and \(T_\text{m}^{(k)}\) represent the total task time on a single GPU and \(k\) GPUs, respectively. 
The speedup rate \(\mu\) for the multi-GPU setup is then defined as:
\begin{equation}
    \mu = \frac{T_\text{s}}{T_\text{m}^{(k)}} = \frac{k\cdot T_\text{s}}{k(k-1)\cdot T_\text{c} + T_\text{s}}.
    \label{eq:rate}
\end{equation}

\begin{wrapfigure}[16]{l}{0.495\textwidth}
    \vskip -2.5em
    \centering
    \includegraphics[width=0.99\linewidth]{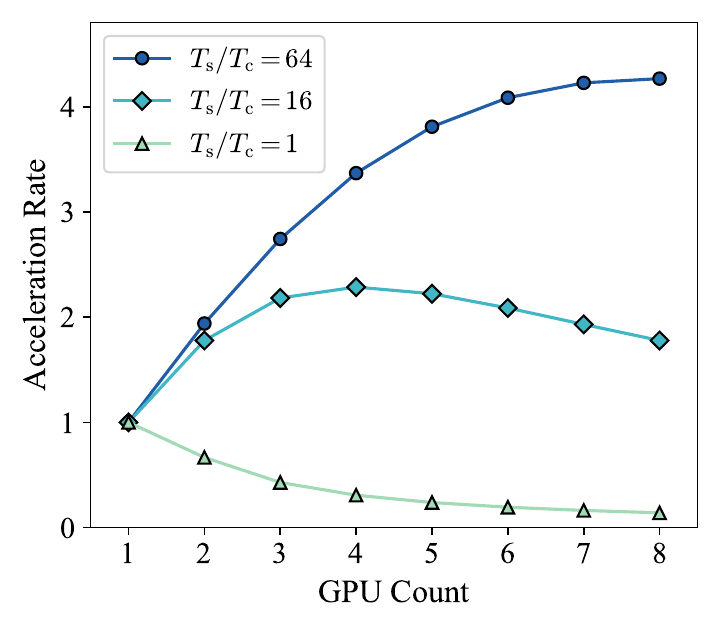}
    \vskip -0.8em
    \caption{The simulated relationship between the acceleration ratio and the GPU count under three task conditions, as per Eq.~\eqref{eq:rate}.}
    \label{fig:rate_curve}
\end{wrapfigure}

Eq.~\eqref{eq:rate} reveals that the acceleration ratio \(\mu\) depends on the GPU count \(k\), influenced by \(T_\text{s}\) and \(T_\text{c}\). 
When considering \(\mu\) as a continuous function of \(k\), its optimum is achieved at \(k = \sqrt{T_\text{s}/T_\text{c}}\). 
Thus, selecting a GPU count \(k\) close to \(\sqrt{T_\text{s}/T_\text{c}}\) maximizes acceleration for given \(T_\text{s}\) and \(T_\text{c}\) values in practical implementations.

Fig.~\ref{fig:rate_curve} demonstrates the simulated relationship between GPU count \(k\) and acceleration ratio \(\mu\) for three distinct \(T_\text{s}/T_\text{c}\) ratios. 
An increase in \(\mu\) is observed as \(k\) nears \(\sqrt{T_\text{s}/T_\text{c}}\). 
Given that \(T_\text{c}\) remains constant across LIF spiking neuron layer time steps, while \(T_\text{s}\) escalates with time step extension, it suggests that multi-GPU deployment becomes increasingly advantageous as SNN time steps expand.

\begin{lstlisting}[label=list:torch-cuda, language=Python, float=!t, belowskip=-2em, caption={Illustration of the programming model for our implementation in PyTorch with CUDA. The \texttt{temporal\_fusion\_kernel} represents the custom CUDA kernel package encapsulating the temporal fusion method. Within this context, \texttt{x} aggregates a tensor for all neurons \(i\) across each time step \(t\), corresponding to \(x_i^{(t)}\) in Eq.~\eqref{eq:LIF}. Similarly, \texttt{grad\_y} denotes the aggregated gradient tensor, corresponding to \(\nabla_{y_i^{(t)}}L\) as delineated in Eq.~\eqref{eq:backLIF}. The \texttt{model} serves as a trainable and testable construct.}]
      import torch
      import temporal_fusion_kernel

      # Define function to acquire kernel
      class FusedLIF(torch.autograd.Function):
        @staticmethod
        def forward(ctx, x, args):
          ctx.args = args
          return temporal_fusion_kernel.fusedForwardLIF(x, args)
        @staticmethod
        def backward(ctx, grad_y):
          args = ctx.args
          return temporal_fusion_kernel.fusedBackwardLIF(grad_y, args)

      # Define LIF layer module
      class LIFLayer(torch.nn.Module):
        def __init__(self, args):
          self.args = args
        def forward(self, x):
          return FusedLIF.apply(x, self.args)
        
      # Define hyper-parameters and model
      args  = ...  # hyper-parameters
      model = torch.nn.Sequential(..., LIFLayer(args), ...)
\end{lstlisting}

\section{Implementation}
This section provides an implementation developed with our temporal fusion method. 
Specifically, the implementation leverages the foundational architecture of PyTorch with CUDA. 
While PyTorch supports the definition and utilization of SNNs, CUDA accommodates both single and multiple GPU operator fusion configurations. 
We begin by introducing the programming models tailored for PyTorch and CUDA based implementation, with a discussion about the implementation schemes and an exploration of the design considerations for the associated functional modules.

To demonstrate the functionalities offered by our implementation, we present the programming model for defining spiking neuron layers with the temporal fusion method, as exemplified in Listing~\ref{list:torch-cuda}.
This example demonstrates how to define a temporally fused LIF neuron layer which can be used to design SNNs optimized for single and multi-GPU acceleration. 

Notably, our implementation primarily focuses on the aspect of designing high-performance spiking neurons. 
To this end, the pre-compilation by CUDA ensures high performance of the proposed temporal fusion, for both forward and backward propagations.
However, modern SNNs usually involve both ANN operators (e.g., convolution) and spiking neurons. 
Hence, from the practical point of view, we integrated our implementation into the PyTorch framework.

\begin{table}[!t]
    \caption{Performance of different implementation methods for training sample SNN architectures on static image datasets and event-based datasets. 
    All time measurements are in seconds. 
    "Acc.", "\(T_\text{train}\)" and "\(T_\text{test}\)" indicate the average of the highest accuracy (in percentage), training time and testing time in 5 epochs, respectively.}
    \label{tab:training-time}
    \setlength{\tabcolsep}{4.2pt}
    \renewcommand{\arraystretch}{1.1}
    \centering
    \begin{tabular}{l|c| c|c|c|c|c|c}
    \hline
    \multirow{3}{*}{\textbf{Architecture}} & \multirow{3}{*}{\textbf{Method}} & \multicolumn{6}{c}{\textbf{Static Image Dataset}} \\
    \cline{3-8} 
    & & \multicolumn{3}{c|}{MNIST} & \multicolumn{3}{c}{CIFAR-10} \\
    \cline{3-8}
    & & Acc. & \(T_\text{train}\) & \(T_\text{test}\) & Acc. & \(T_\text{train}\) & \(T_\text{test}\) \\
    \hline
    \multirow{3}{*}{\makecell[c]{Spiking-ResNet18}}
    & SpikingJelly\textsuperscript{a} & 98.18 & 170.31 & 10.52 & 55.94 & 160.59 & 12.97 \\
    \cline{2-8} 
    & SpikingJelly\textsuperscript{b} & 98.35 & 70.91 & 11.13 & 54.54 & 71.17 & 13.00 \\
    \cline{2-8} 
    & \textbf{Ours} & 98.29 & \textbf{65.64} & \textbf{9.12} & 54.33 & \textbf{68.11} & \textbf{11.56} \\
    \hline
    \multirow{3}{*}{\makecell[c]{Spiking-ResNet34}}
    & SpikingJelly\textsuperscript{a} & 96.98 & 301.21 & 12.81 & 30.80 & 265.94 & 15.96 \\
    \cline{2-8} 
    & SpikingJelly\textsuperscript{b} & 96.53 & 90.41 & 13.58 & 37.25 & 86.21 & 15.71 \\
    \cline{2-8} 
    & \textbf{Ours} & 96.53 & \textbf{81.56} & \textbf{10.67} & 33.16 & \textbf{81.90} & \textbf{12.99} \\
    \hline
    \multirow{3}{*}{\makecell[c]{Spiking-ResNet50}}
    & SpikingJelly\textsuperscript{a} & 95.64 & 455.81 & 19.28 & 27.03 & 413.58 & 21.62 \\
    \cline{2-8} 
    & SpikingJelly\textsuperscript{b} & 96.62 & 154.09 & 19.29 & 23.39 & 141.14 & 21.66 \\
    \cline{2-8} 
    & \textbf{Ours} & 96.10 & \textbf{144.69} & \textbf{16.44} & 29.48 & \textbf{131.78} & \textbf{18.76} \\
    \hline
    
    \multirow{3}{*}{\textbf{Architecture}} & \multirow{3}{*}{\textbf{Method}} & \multicolumn{6}{c}{\textbf{Event-Based Dataset}} \\
    \cline{3-8} 
    & & \multicolumn{3}{c|}{N-MNIST} & \multicolumn{3}{c}{DvsGesture} \\
    \cline{3-8}
    & & Acc. & \(T_\text{train}\) & \(T_\text{test}\) & Acc. & \(T_\text{train}\) & \(T_\text{test}\) \\
    \hline
    \multirow{3}{*}{\makecell[c]{Spiking-ResNet18}}
    & SpikingJelly\textsuperscript{a} & 98.24 & 251.14 & 21.44 & 64.64 & 55.61 & 7.10 \\
    \cline{2-8} 
    & SpikingJelly\textsuperscript{b} & 98.39 & 142.33 & 21.83 & 65.71 & 31.46 & 7.37 \\
    \cline{2-8} 
    & \textbf{Ours} & 97.13 & \textbf{140.04} & \textbf{20.02} & 66.67 & \textbf{29.40} & \textbf{6.10} \\
    \hline
    \multirow{3}{*}{\makecell[c]{Spiking-ResNet34}}
    & SpikingJelly\textsuperscript{a} & 11.35 & 374.04 & 23.97 & 52.98 & 80.80 & 7.78 \\
    \cline{2-8} 
    & SpikingJelly\textsuperscript{b} & 11.35 & 170.90 & 24.33 & 55.60 & 34.32 & 7.80 \\
    \cline{2-8} 
    & \textbf{Ours} & 11.35 & \textbf{165.61} & \textbf{21.76} & 56.48 & \textbf{31.92} & \textbf{6.26} \\
    \hline
    \multirow{3}{*}{\makecell[c]{Spiking-ResNet50}}
    & SpikingJelly\textsuperscript{a} & 10.59 & 626.19 & 32.43 & 52.14 & 119.38 & 8.68 \\
    \cline{2-8} 
    & SpikingJelly\textsuperscript{b} & 10.59 & 255.58 & 32.42 & 57.62 & 44.76 & 8.83 \\
    \cline{2-8} 
    & \textbf{Ours} & 11.01 & \textbf{224.83} & \textbf{28.76} & 55.09 & \textbf{38.64} & \textbf{6.80} \\
    \hline
    
    \multicolumn{8}{l}{\textsuperscript{a} On PyTorch back end.} \\
    \multicolumn{8}{l}{\textsuperscript{b} On CuPy back end.} \\
    \end{tabular}
    \renewcommand{\arraystretch}{1}
    \vskip -1em
\end{table}

\section{Experiment}
Our experiment comprises three parts.
Initially, we evaluated the temporal fusion method within various SNN architectures across a spectrum of time steps, utilizing both static image and event-based datasets. 
Subsequent experiments focused on assessing the temporal fusion method's performance using a LIF unit across different time steps, highlighting optimal conditions. 
Lastly, we present benchmark results from multi-GPU configurations, emphasizing the scalability and advantages of employing multiple GPUs.

\begin{figure}[!ht]
  \centering
  \begin{minipage}{0.655\textwidth}
    \centering
    \includegraphics[width=0.95\linewidth]{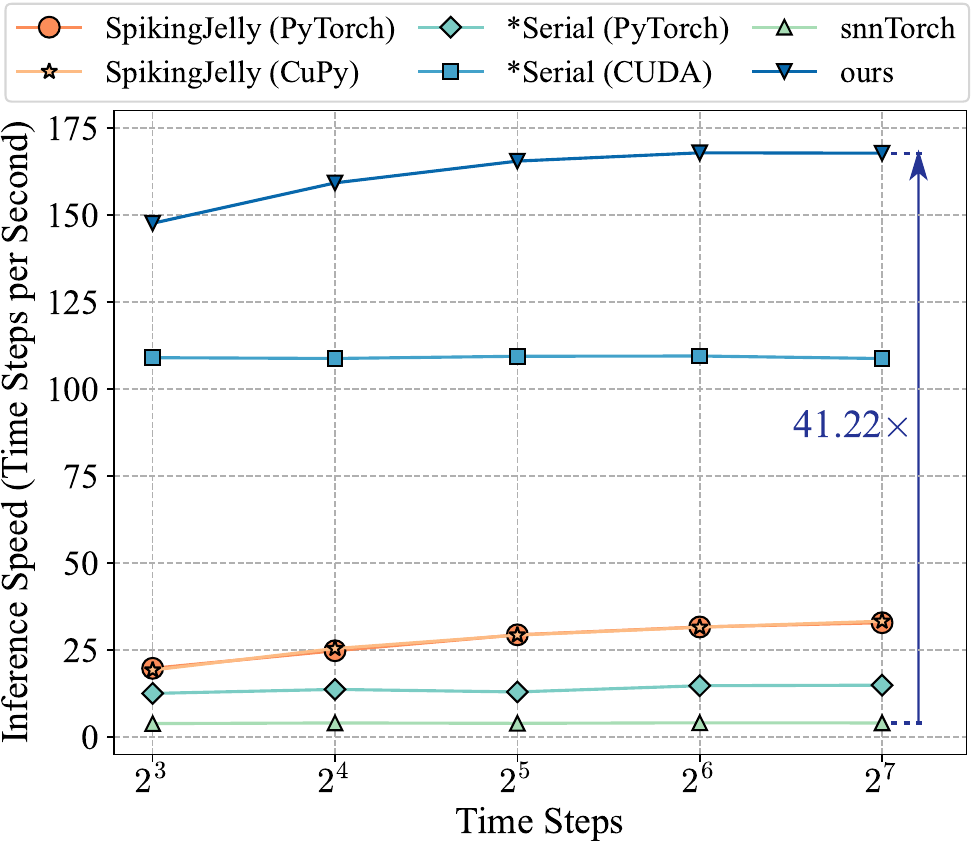}
  \end{minipage}
  \begin{minipage}{0.335\textwidth}
    \caption{Performance comparison of multiple methods for training monolayer LIF on single GPU at different time steps. The * symbols mark the self-implemented baselines, "Serial (PyTorch)" and "Serial (CUDA)" correspond to the PyTorch-based and the CUDA-based implementations of "Serial Training" method (see Fig.~\ref{fig:fused_analysis}), respectively.}
    \label{fig:single_acceleration}
  \end{minipage}
  \vskip -1.5em
\end{figure}

\subsection{Experimental Setup}
All experiments were meticulously carried out on NVIDIA A100 GPUs to encompass runtime comparisons across different platforms and implementations.
To ensure equitable comparisons, we aligned the hyperparameters as specified in Eq.~\eqref{eq:LIF}, setting \(V_\text{rest}=0\), \(k_\tau=0.2\), and \(V_\text{th}=0.3\). 
For libraries lacking the \(k_\tau\) hyperparameter, we substituted with \(\tau=1.25\) based on the relation \(k_{\tau} = 1 - 1/\tau\).

\subsection{Main Results}

In this experiment, we selected three architectures: Spiking-ResNet18, Spiking-ResNet34, and Spiking-ResNet50, as informed by existing research~\cite{doi:10.1126/sciadv.adi1480,DBLP:conf/cvpr/HeZRS16}. Referring to the SpikingJelly, we integrated these architectures with our acceleration method for comprehensive testing. 
The learning rate was set to \(0.001\) in the Adam optimizer~\cite{DBLP:journals/corr/KingmaB14}, with time steps fixed at \(32\) over \(5\) epochs, and the employment of cross-entropy loss. 
We adopted the sigmoid surrogate function as introduced in~\cite{10.3389/fnins.2018.00331}:
\begin{equation}
    \delta(x) = \sigma'(x) = \frac{\alpha e^{-\alpha x}}{(1+e^{-\alpha x})^2},
\end{equation}
with \(\alpha=4.0\). 
All tests across various datasets and architectures were conducted three times, averaging the results to ensure reliability.

Initially, we employed two static image datasets, MNIST~\cite{DBLP:journals/pieee/LeCunBBH98} and CIFAR-10~\cite{krizhevsky2009learning}, to evaluate both the acceleration capabilities and the accuracy of our method. 
For these static datasets, the Poisson Sampling method was used during pre-processing, with a batch size of 100. Subsequently, we applied the same architectures to two event-based datasets, N-MNIST~\cite{10.3389/fnins.2015.00437} and DvsGesture~\cite{DBLP:conf/cvpr/AmirTBMMNNAGMKD17}, adjusting batch sizes to 100 and 10, respectively.
During pre-processing for these datasets, we utilized the Stacking Based on the number of Events (SBE) approach as implemented in SpikingJelly.

As summarized in Table~\ref{tab:training-time}, the results indicate that our method achieves comparable accuracy to existing implementations across different network architectures for both static and dynamic datasets, while also ensuring considerable speedups in both training and testing phases.

\subsection{Scalable Time Steps}
This experiment assesses our method's scalability with respect to time steps in a single-GPU setting. 
To focus solely on our method's acceleration performance, all ANN operators were omitted, centering the experiment on a single-layer LIF model with 1,000,000 neurons. 
Results reflect the aggregated time from 1,000 independent trials.

Fig.~\ref{fig:single_acceleration} showcases our temporal fusion method's remarkable acceleration performance, achieving up to a \(40\times\) speedup over traditional implementations, demonstrating its scalability concerning time steps.

\subsection{Multi-GPU Acceleration}
\begin{figure}[!t]
    \begin{center}
    \includegraphics[width=1\textwidth]{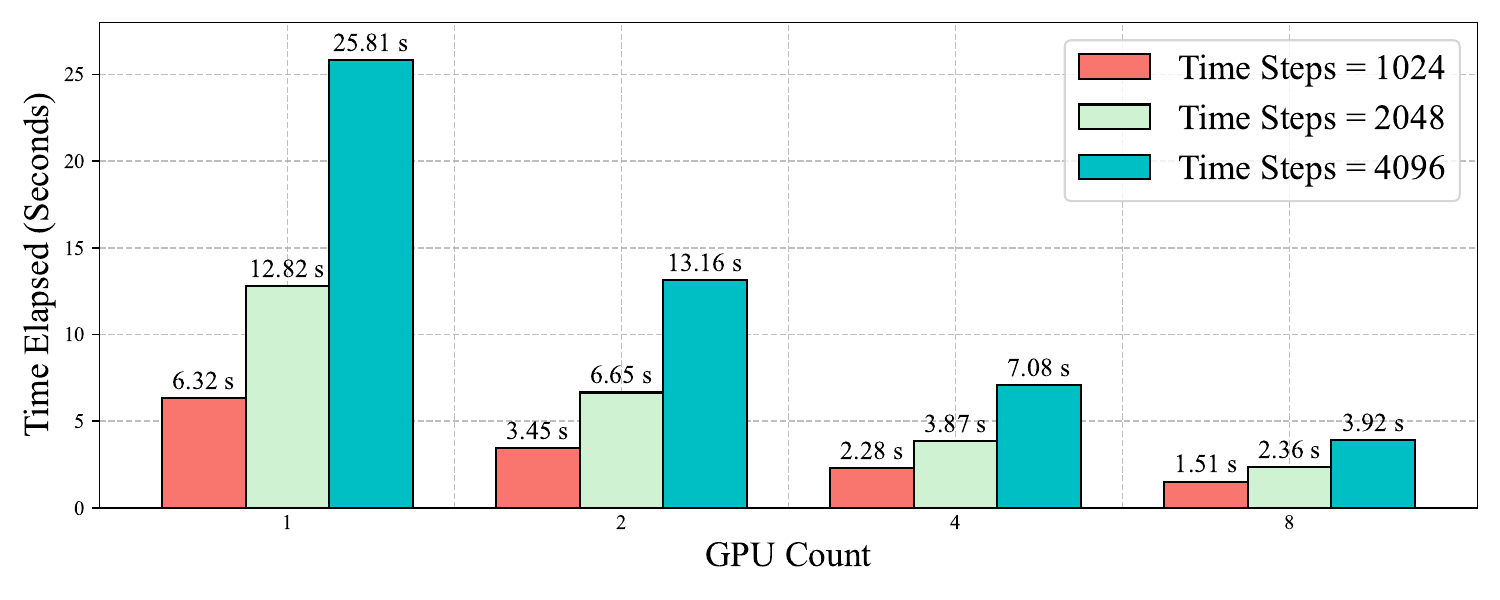}
    \end{center}
    \vskip -2em
    \caption{Performance of multi-GPU acceleration using the CUDA implementation of our temporal fusion method. }
    \label{fig:acceleration}
    \vskip -1em
\end{figure}

The multi-GPU experiment involved deploying the temporal fusion method across 1 to 8 NVIDIA A100 GPUs, focusing on a monolayer LIF model with 1,000,000 neurons across various time steps. 
The reported outcomes are averages over 1,000 independent runs.
As shown in Fig.~\ref{fig:acceleration}, a notable decrease in computational time was observed with increasing GPU count, highlighting our method's robust performance in a multi-GPU setting.

\section{Conclusion}
In this paper, we present a method, termed "temporal fusion", specifically designed for the efficient GPU-accelerated training of SNNs. To facilitate the practical application of this method, we have developed a CUDA-based implementation, complete with a detailed programming model. This implementation is seamlessly integrated with the widely-used deep learning framework, PyTorch, thereby enabling users to easily adopt and apply our method in their SNN research and development projects.
We have conducted a series of comprehensive benchmark tests on the temporal fusion method. These tests are meticulously designed to assess various aspects of the method, including its adaptability to different SNN architectures and training scenarios, as well as its impact on the overall efficiency of the training process. The results demonstrate the significant improvements in the training efficiency and wide applicability of our method.

While temporal fusion presents a promising avenue for scaling SNNs, certain challenges persist in our current framework. 
For instance, although our experimental findings confirm the viability of temporal fusion across both single and multi-GPU setups, the scalability of SNNs to larger dimensions may surpass the capacity of multi-GPU configurations within a single node. 
As such, extending support to multi-node systems for large-scale SNN acceleration emerges as an imminent challenge, awaiting resolution.
Moreover, the proposed solution is more focused on the forward and backward propagation acceleration of traditional deep learning-based training for SNNs. Despite the ongoing importance of this training method, it might lead to the loss of some temporal information, research on acceleration methods closer to the spiking neuron mechanism is one of the future works.

Since SNN research has started to focus more on larger temporal dimensions as they possess better temporal dynamics, we endeavour to research better utilization of these characteristics in future via temporal fusion. 
Importantly, harnessing the capabilities of large-scale SNNs could pave the way for advancements that edge closer to true intelligence, linking directly to the investigation of foundational support for large-scale models. 
The exploration of how to effectively synergize these two facets holds substantial promise as a future research direction.

\bibliographystyle{splncs04}
\bibliography{main}

\begin{thebibliography}{10}
\providecommand{\url}[1]{\texttt{#1}}
\providecommand{\urlprefix}{URL }
\providecommand{\doi}[1]{https://doi.org/#1}

\bibitem{DBLP:conf/cvpr/AmirTBMMNNAGMKD17}
Amir, A., Taba, B., Berg, D.J., Melano, T., McKinstry, J.L., di~Nolfo, C., Nayak, T.K., Andreopoulos, A., Garreau, G., Mendoza, M., Kusnitz, J., DeBole, M., Esser, S.K., Delbr{\"{u}}ck, T., Flickner, M., Modha, D.S.: A low power, fully event-based gesture recognition system. In: 2017 {IEEE} Conference on Computer Vision and Pattern Recognition, {CVPR}. pp. 7388--7397. {IEEE} Computer Society (2017)

\bibitem{DBLP:conf/iclr/BuFDDY022}
Bu, T., Fang, W., Ding, J., Dai, P., Yu, Z., Huang, T.: Optimal {ANN-SNN} conversion for high-accuracy and ultra-low-latency spiking neural networks. In: The Tenth International Conference on Learning Representations, {ICLR}. OpenReview.net (2022)

\bibitem{DBLP:journals/corr/ChetlurWVCTCS14}
Chetlur, S., Woolley, C., Vandermersch, P., Cohen, J., Tran, J., Catanzaro, B., Shelhamer, E.: cu{DNN}: Efficient primitives for deep learning. CoRR  \textbf{abs/1410.0759} (2014)

\bibitem{DBLP:journals/ijon/ChowdhuryLR21}
Chowdhury, S.S., Lee, C., Roy, K.: Towards understanding the effect of leak in spiking neural networks. Neurocomputing  \textbf{464},  83--94 (2021)

\bibitem{DBLP:conf/eccv/ChowdhuryRR22}
Chowdhury, S.S., Rathi, N., Roy, K.: Towards ultra low latency spiking neural networks for vision and sequential tasks using temporal pruning. In: Computer Vision --- {ECCV}. Lecture Notes in Computer Science, vol. 13671, pp. 709--726. Springer (2022)

\bibitem{DBLP:conf/ijcai/DingY0H21}
Ding, J., Yu, Z., Tian, Y., Huang, T.: Optimal {ANN-SNN} conversion for fast and accurate inference in deep spiking neural networks. In: Proceedings of the Thirtieth International Joint Conference on Artificial Intelligence, {IJCAI}. pp. 2328--2336 (2021)

\bibitem{DBLP:journals/pieee/EshraghianWNWLDBJL23}
Eshraghian, J.K., Ward, M., Neftci, E.O., Wang, X., Lenz, G., Dwivedi, G., Bennamoun, M., Jeong, D.S., Lu, W.D.: Training spiking neural networks using lessons from deep learning. Proc. {IEEE}  \textbf{111}(9),  1016--1054 (2023)

\bibitem{doi:10.1126/sciadv.adi1480}
Fang, W., Chen, Y., Ding, J., Yu, Z., Masquelier, T., Chen, D., Huang, L., Zhou, H., Li, G., Tian, Y.: Spiking{J}elly: An open-source machine learning infrastructure platform for spike-based intelligence. Science Advances  \textbf{9}(40),  eadi1480 (2023)

\bibitem{fang2023parallel}
Fang, W., Yu, Z., Zhou, Z., Chen, D., Chen, Y., Ma, Z., Masquelier, T., Tian, Y.: Parallel spiking neurons with high efficiency and ability to learn long-term dependencies. In: Thirty-seventh Conference on Neural Information Processing Systems (2023)

\bibitem{gerstner_kistler_naud_paninski_2014}
Gerstner, W., Kistler, W.M., Naud, R., Paninski, L.: Neuronal dynamics: From single neurons to networks and models of cognition. Cambridge University Press (2014)

\bibitem{11.3389/fnins.2023.1047008}
Guo, W., Fouda, M.E., Eltawil, A.M., Salama, K.N.: Efficient training of spiking neural networks with temporally-truncated local backpropagation through time. Frontiers in Neuroscience  \textbf{17} (2023)

\bibitem{DBLP:conf/cvpr/0006S020}
Han, B., Srinivasan, G., Roy, K.: {RMP-SNN:} {Residual} membrane potential neuron for enabling deeper high-accuracy and low-latency spiking neural network. In: 2020 {IEEE/CVF} Conference on Computer Vision and Pattern Recognition, {CVPR}. pp. 13555--13564. Computer Vision Foundation / {IEEE} (2020)

\bibitem{DBLP:journals/fini/HazanSKPSSK18}
Hazan, H., Saunders, D.J., Khan, H., Patel, D., Sanghavi, D.T., Siegelmann, H.T., Kozma, R.: Binds{NET}: {A} machine learning-oriented spiking neural networks library in {P}ython. Frontiers Neuroinformatics  \textbf{12}, ~89 (2018)

\bibitem{DBLP:conf/cvpr/HeZRS16}
He, K., Zhang, X., Ren, S., Sun, J.: Deep residual learning for image recognition. In: 2016 {IEEE} Conference on Computer Vision and Pattern Recognition, {CVPR}. pp. 770--778. {IEEE} Computer Society (2016)

\bibitem{https://doi.org/10.1113/jphysiol.1952.sp004717}
Hodgkin, A.L., Huxley, A.F.: Currents carried by sodium and potassium ions through the membrane of the giant axon of {Loligo}. The Journal of Physiology  \textbf{116}(4),  449--472 (1952)

\bibitem{DBLP:conf/nips/HuangCBFCCLNLWC19}
Huang, Y., Cheng, Y., Bapna, A., Firat, O., Chen, D., Chen, M.X., Lee, H., Ngiam, J., Le, Q.V., Wu, Y., Chen, Z.: {GPipe}: Efficient training of giant neural networks using pipeline parallelism. In: Advances in Neural Information Processing Systems, NeurIPS. pp. 103--112 (2019)

\bibitem{Izhikevich20031569}
Izhikevich, E.M.: Simple model of spiking neurons. IEEE Transactions on Neural Networks  \textbf{14}(6),  1569--1572 (2003)

\bibitem{DBLP:journals/corr/KingmaB14}
Kingma, D.P., Ba, J.: Adam: {A} method for stochastic optimization. In: 3rd International Conference on Learning Representations, {ICLR} (2015)

\bibitem{krizhevsky2009learning}
Krizhevsky, A., Hinton, G., et~al.: Learning multiple layers of features from tiny images  (2009)

\bibitem{DBLP:conf/nips/KrizhevskySH12}
Krizhevsky, A., Sutskever, I., Hinton, G.E.: {ImageNet} classification with deep convolutional neural networks. In: Advances in Neural Information Processing Systems. pp. 1106--1114 (2012)

\bibitem{DBLP:journals/pieee/LeCunBBH98}
LeCun, Y., Bottou, L., Bengio, Y., Haffner, P.: Gradient-based learning applied to document recognition. Proc. {IEEE}  \textbf{86}(11),  2278--2324 (1998)

\bibitem{DBLP:conf/icml/LiD0GG21}
Li, Y., Deng, S., Dong, X., Gong, R., Gu, S.: A free lunch from {ANN}: Towards efficient, accurate spiking neural networks calibration. In: Proceedings of the 38th International Conference on Machine Learning, {ICML}. Proceedings of Machine Learning Research, vol.~139, pp. 6316--6325 (2021)

\bibitem{DBLP:conf/nips/LiGZDHG21}
Li, Y., Guo, Y., Zhang, S., Deng, S., Hai, Y., Gu, S.: Differentiable spike: Rethinking gradient-descent for training spiking neural networks. In: Advances in Neural Information Processing Systems, NeurIPS. pp. 23426--23439 (2021)

\bibitem{DBLP:conf/isbi/Luebke08}
Luebke, D.P.: {CUDA:} {Scalable} parallel programming for high-performance scientific computing. In: Proceedings of the 2008 {IEEE} International Symposium on Biomedical Imaging: From Nano to Macro. pp. 836--838. {IEEE} (2008)

\bibitem{DBLP:conf/iccv/Meng0Y0LL23}
Meng, Q., Xiao, M., Yan, S., Wang, Y., Lin, Z., Luo, Z.: Towards memory- and time-efficient backpropagation for training spiking neural networks. In: {IEEE/CVF} International Conference on Computer Vision, {ICCV}. pp. 6143--6153. {IEEE} (2023)

\bibitem{10.3389/fnins.2015.00437}
Orchard, G., Jayawant, A., Cohen, G.K., Thakor, N.: Converting static image datasets to spiking neuromorphic datasets using saccades. Frontiers in Neuroscience  \textbf{9} (2015)

\bibitem{DBLP:journals/pami/Paredes-VallesS20}
Paredes{-}Vall{\'{e}}s, F., Scheper, K.Y.W., de~Croon, G.C.H.E.: Unsupervised learning of a hierarchical spiking neural network for optical flow estimation: From events to global motion perception. {IEEE} Trans. Pattern Anal. Mach. Intell.  \textbf{42}(8),  2051--2064 (2020)

\bibitem{DBLP:conf/nips/PaszkeGMLBCKLGA19}
Paszke, A., Gross, S., Massa, F., Lerer, A., Bradbury, J., Chanan, G., Killeen, T., Lin, Z., Gimelshein, N., Antiga, L., Desmaison, A., K{\"{o}}pf, A., Yang, E.Z., DeVito, Z., Raison, M., Tejani, A., Chilamkurthy, S., Steiner, B., Fang, L., Bai, J., Chintala, S.: Py{T}orch: An imperative style, high-performance deep learning library. In: Advances in Neural Information Processing Systems, NeurIPS. pp. 8024--8035 (2019)

\bibitem{norse2021}
Pehle, C., Pedersen, J.E.: {Norse --- A deep learning library for spiking neural networks} (Jan 2021)

\bibitem{DBLP:journals/tnn/RathiR23}
Rathi, N., Roy, K.: {DIET-SNN:} {A} low-latency spiking neural network with direct input encoding and leakage and threshold optimization. {IEEE} Trans. Neural Networks Learn. Syst.  \textbf{34}(6),  3174--3182 (2023)

\bibitem{10.3389/fnins.2017.00682}
Rueckauer, B., Lungu, I.A., Hu, Y., Pfeiffer, M., Liu, S.C.: Conversion of continuous-valued deep networks to efficient event-driven networks for image classification. Frontiers in Neuroscience  \textbf{11} (2017)

\bibitem{shaban2021adaptive}
Shaban, A., Bezugam, S.S., Suri, M.: An adaptive threshold neuron for recurrent spiking neural networks with nanodevice hardware implementation. Nature Communications  \textbf{12}(1), ~4234 (2021)

\bibitem{DBLP:journals/corr/abs-1909-08053}
Shoeybi, M., Patwary, M., Puri, R., LeGresley, P., Casper, J., Catanzaro, B.: {Megatron-LM}: Training multi-billion parameter language models using model parallelism. CoRR  \textbf{abs/1909.08053} (2019)

\bibitem{10.1007/978-3-030-30487-4_54}
State, L., Vilimelis~Aceituno, P.: Training delays in spiking neural networks. In: Artificial Neural Networks and Machine Learning --- ICANN. pp. 713--717. Springer International Publishing, Cham (2019)

\bibitem{teeter2018generalized}
Teeter, C., Iyer, R., Menon, V., Gouwens, N., Feng, D., Berg, J., Szafer, A., Cain, N., Zeng, H., Hawrylycz, M., et~al.: Generalized leaky integrate-and-fire models classify multiple neuron types. Nature Communications  \textbf{9}(1), ~709 (2018)

\bibitem{10.3389/fnins.2018.00331}
Wu, Y., Deng, L., Li, G., Zhu, J., Shi, L.: Spatio-temporal backpropagation for training high-performance spiking neural networks. Frontiers in Neuroscience  \textbf{12} (2018)

\bibitem{DBLP:conf/aaai/WuDLZ0S19}
Wu, Y., Deng, L., Li, G., Zhu, J., Xie, Y., Shi, L.: Direct training for spiking neural networks: Faster, larger, better. In: The Thirty-Third {AAAI} Conference on Artificial Intelligence, {AAAI} 2019, The Thirty-First Innovative Applications of Artificial Intelligence Conference, {IAAI}. pp. 1311--1318. {AAAI} Press (2019)

\bibitem{DBLP:conf/nips/XiaoMZHL22}
Xiao, M., Meng, Q., Zhang, Z., He, D., Lin, Z.: Online training through time for spiking neural networks. In: Advances in Neural Information Processing Systems, NeurIPS (2022)

\bibitem{yavuz2016genn}
Yavuz, E., Turner, J., Nowotny, T.: Ge{NN}: {A} code generation framework for accelerated brain simulations. Scientific Reports  \textbf{6},  18854 (2016)

\bibitem{10.1007/978-3-031-44192-9_33}
Zhang, Y., Cao, J., Chen, J., Sun, W., Wang, Y.: {Razor SNN}: Efficient spiking neural network with temporal embeddings. In: Artificial Neural Networks and Machine Learning --- ICANN. pp. 411--422. Springer Nature Switzerland, Cham (2023)

\end{thebibliography}

\end{document}